\pdfoutput=1

\documentclass[11pt]{article}

\usepackage[final]{acl}

\usepackage{times}
\usepackage{latexsym}

\usepackage[T1]{fontenc}

\usepackage[utf8]{inputenc}

\usepackage{microtype}

\usepackage{inconsolata}

\usepackage{amsmath}
\usepackage{times}
\usepackage{framed}
\usepackage{ragged2e}
\usepackage{bm}
\usepackage{soul}
\usepackage{latexsym}
\usepackage{subfigure}
\usepackage{amsthm}
\usepackage{graphicx}
\usepackage{float}
\usepackage{longtable}
\usepackage{booktabs} 
\usepackage{multirow}
\usepackage{multicol}
\usepackage{amssymb}
\usepackage{dashbox}%
\usepackage{multirow}
\usepackage{textcomp}
\usepackage{tcolorbox}
\usepackage{bbding}
\usepackage{bbm}
\usepackage[ruled, vlined, linesnumbered]{algorithm2e}
\usepackage{mathtools}
\usepackage{adjustbox}
\usepackage[ruled,vlined]{algorithm2e}
\usepackage{color, soul}
\usepackage{pifont}
\usepackage{array}
\newcolumntype{P}[1]{>{\centering\arraybackslash}p{#1}}
\newcolumntype{M}[1]{>{\centering\arraybackslash}m{#1}}
\usepackage{cleveref}
\crefname{section}{§}{§§}
\Crefname{section}{§}{§§}
\crefname{figure}{Figure}{Figure}
\Crefname{figure}{Figure}{Figure}
\crefname{table}{Table}{Table}
\Crefname{table}{Table}{Table}
\usepackage{tabularx}
\newcommand\ourmodel{ADELIE\textsubscript{SFT}\xspace}
\newcommand\ourmodeldpo{ADELIE\textsubscript{DPO}\xspace}
\newcommand\ourdata{\texttt{IEInstruct}\xspace}
\newcommand\ourdatadpo{\texttt{IEFeedback}\xspace}

\title{ADELIE: Aligning Large Language Models on \\Information Extraction}

\author{Yunjia Qi\thanks{Equal contribution.}, Hao Peng \hspace{-3pt}$^*$, Xiaozhi Wang, Bin Xu\thanks{Corresponding author: xubin@tsinghua.edu.cn}, Lei Hou, Juanzi Li \\
        Department of Computer Science and Technology, BNRist, Tsinghua University \\ \texttt{\{qyj23, peng-h24\}@mails.tsinghua.edu.cn}}

\begin{document}
\maketitle
\begin{abstract}

Large language models (LLMs) usually fall short on information extraction (IE) tasks and struggle to follow the complex instructions of IE tasks. This primarily arises from LLMs not being aligned with humans, as mainstream alignment datasets typically do not include IE data.
In this paper, we introduce \textbf{ADELIE} (\textbf{A}ligning large language mo\textbf{DEL}s on \textbf{I}nformation \textbf{E}xtraction), an aligned LLM that effectively solves various IE tasks, including closed IE, open IE, and on-demand IE. We first collect and construct a high-quality alignment corpus \ourdata for IE. Then we train \ourmodel using instruction tuning on \ourdata. We further train \ourmodel with direct preference optimization (DPO) objective, resulting in \ourmodeldpo. Extensive experiments on various held-out IE datasets demonstrate that our models (\ourmodel and \ourmodeldpo) achieve state-of-the-art (SoTA) performance among open-source models. We further explore the general capabilities of ADELIE, and experimental results reveal that their general capabilities do not exhibit a noticeable decline. We have released the code, data, and models to facilitate further research.\footnote{https://github.com/THU-KEG/ADELIE}

\end{abstract}

\section{Introduction}


Large language models (LLMs), especially after alignment with human expectations, such as instruction tuning~\citep{wei2021finetuned, chung2022scaling, longpre2023flan} or direct preference optimization (DPO)~\citep{dpo}, have achieved impressive results on numerous tasks~\citep{chatgpt, openai2023gpt, mistral-7b, gemini, claude-3}. However, LLMs still fall short on information extraction (IE) tasks, particularly on closed IE tasks~\citep{li2023evaluating, han2023information, peng2023specification}. 
LLMs usually struggle to understand and follow the complex instructions of IE tasks~\citep{peng2023specification, pang2023guideline, Xu2023LargeLM}, e.g., complicated task schema and specifications, which indicates existing LLMs are not aligned with human needs on IE tasks~\citep{peng2023specification, Sainz2023GoLLIEAG}.
\begin{figure}[t]
    \includegraphics[width=1.0\linewidth]{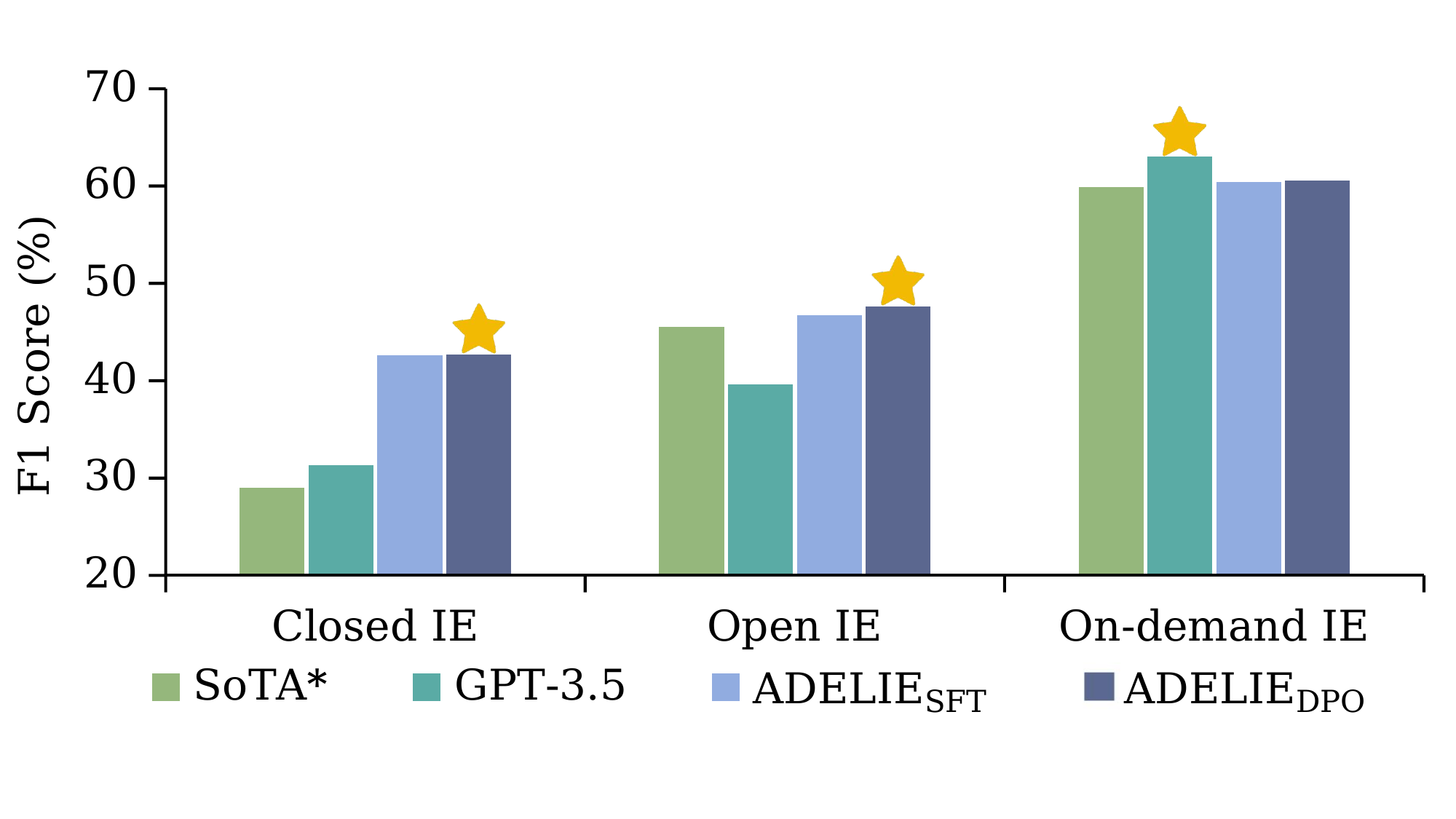} 
    \caption{
        F1 scores (\%) on closed, open, and on-demand IE tasks in the few-shot setting. SoTA* denotes the best performance of open-source models. 
    }
    \label{fig:generalized IE overall score}
\end{figure}

To enhance LLM performance on IE tasks, existing efforts have primarily explored three aspects: (1) \textbf{Prompt engineering}, which provides comprehensive information, e.g., annotation guidelines, to LLMs, without fine-tuning model parameters~\citep{pang2023guideline, Guo2023RetrievalAugmentedCG, Wei2023ZeroShotIE,wan2023gpt}. (2) \textbf{Code LLMs}, which leverage their capabilities of understanding structured information to enhance the performance on IE tasks~\citep{Guo2023RetrievalAugmentedCG, Sainz2023GoLLIEAG, Bi2023CodeKGCCL}.
(3) \textbf{Multi-task fine-tuning}, which involves fine-tuning LLMs on multiple IE datasets to enhance their cross-task generalization capabilities in solving IE tasks~\citep{Wang2022DeepStruct, Wang2023InstructUIEMI, Sainz2023GoLLIEAG, wang2023code4struct}.

However, these works do not sufficiently align LLMs on IE tasks. The prompt engineering method does not inherently align LLMs without tuning model parameters. Works using code LLMs and multi-task fine-tuning typically fine-tune models on homogeneous data, e.g., instances with the same input-output format, with a lack of diverse alignment data.
Therefore, the fine-tuned models exhibit limited generalization capabilities on IE tasks, including closed IE~\citep{Xu2023LargeLM}, open IE~\citep{Xu2023LargeLM}, and on-demand IE~\citep{Jiao2023InstructAE}. Furthermore, as these models are trained specifically for IE, their general capabilities, such as natural language understanding~\citep{Hendrycks2020MeasuringMM}, may experience a significant decline.

Considering the above issues, we introduce \textbf{ADELIE} (\textbf{A}ligning large language mo\textbf{DEL}s on \textbf{I}nformation \textbf{E}xtraction), an LLM aligned on IE tasks. Specifically, this work addresses the above limitations through two aspects: (1) \textbf{Rich alignment data}. We construct a high-quality instruction tuning dataset for IE tasks, \ourdata, including $83,585$ instances of various IE tasks. \ourdata includes a diverse set of instructions and input-output formats. We manually write several instructions for different IE tasks, then expand the instruction set using GPT-3.5~\citep{chatgpt} similar to Self-Instruct~\citep{wang2023self}. We then augment the instructions through various augmentation techniques, such as adding annotation guidelines~\citep{Sainz2023GoLLIEAG}. \ourdata also includes diverse output formats, such as triplets, natural language, and JSON. We also employ GPT-4~\citep{openai2023gpt} to generate chain-of-thought explanations~\citep{wei2022chain} for $8,000$ instances in \ourdata. (2) \textbf{Sufficient alignment}. \ourmodel is trained based on LLAMA 2~\citep{touvron2023llama}, using supervised fine-tuning (SFT)~\citep{ouyang2022training} on a mixture of \ourdata and generic alignment data used in TULU 2~\citep{Ivison2023CamelsIA} to maintain the model's general capabilities. We further train \ourmodel using the direct preference optimization (DPO) objective~\citep{dpo} on \ourdatadpo, a preference dataset constructed using \ourmodel, resulting in \ourmodeldpo.

We comprehensively evaluate \ourmodel and \ourmodeldpo on closed, open, and on-demand IE. Some results are shown in Figure~\ref{fig:generalized IE overall score}.
The results demonstrate that our models achieve SoTA performance compared to previous open-source models and GPT-3.5. 
There is no significant decline in ADELIE's general capabilities, such as MMLU~\citep{Hendrycks2020MeasuringMM} and BBH~\citep{suzgun2023challenging}. Moreover, we analyze several key factors of the alignment process and provide several insightful findings, such as the mixture strategy of IE and general alignment data.
We hope our extensive experiments and analyses will advance research on aligning LLMs.

In summary, our contributions are threefold: (1) We construct high-quality alignment data for IE tasks: \ourdata and \ourdatadpo. (2) Based on this high-quality alignment data, we develop \ourmodel and \ourmodeldpo, with advanced performance on IE tasks. (3) We conduct extensive experiments and analyses, providing meaningful insights for the research on LLM alignment.



\section{Related Work}

\subsection{Information Extraction Tasks}
Conventional IE tasks are primarily categorized into two types: closed IE and open IE. Closed IE involves extracting structured information from unstructured text, typically requiring the extracted information to conform to a predefined schema. Closed IE includes the following tasks:
(1) Named Entity Recognition (NER), which aims to identify entities in text and categorizing them into types defined in a schema~\citep{yadav2018survey}.
(2) Relation Classification (RC), which classifies the relationship into a predefined type between two mentioned entities in the text~\citep{han2020more}.
(3) Relation Extraction (RE), which aims to extract entities and their relations end-to-end~\citep{zhong2021frustratingly}.
(4) Event Detection (ED), which extracts event triggers and classifies them into predefined types~\cite{wang2020maven}.
(5) Event Argument Extraction (EAE), which aims to extract arguments, e.g., time, for events~\cite{wang2023maven}.
(6) Event Extraction (EE), which aims to extract events and their arguments in end-to-end paradigm~\citep{peng2023devil}.
(7) Event Relation Extraction (ERE), which extracts coreference, temporal, causal, and hierarchical relationships between events~\cite{wang2022maven}. 
Open IE aims to extract n-ary relation tuples from text, without relying on a predefined schema~\citep{zhousurvey}.

Beyond closed IE and open IE, \citet{Jiao2023InstructAE} proposed on-demand IE, aimed at extracting user-desired information from unstructured text, such as extracting the shape and taste of fruits, and organizing it into a structured tabular format. On-demand IE is more flexible and aligns with real-world user demand. This paper covers all these IE tasks, aiming to enhance the model's ability to address these tasks through sufficient alignment.

\subsection{LLMs for Information Extraction}
LLMs often fall short on IE tasks~\citep{li2023evaluating, han2023information} due to the complex specifications of these tasks~\citep{peng2023specification}. Consequently, numerous works have been proposed to enhance LLMs' understanding of IE task specifications to improve their performance. These works are primarily divided into three aspects: (1) Prompt engineering~\citep{pang2023guideline, Guo2023RetrievalAugmentedCG, Wei2023ZeroShotIE,wang2023gpt,wan2023gpt,zhang20232iner,xie2023self}, aims to enhance the model's performance on IE tasks by providing sufficient prompts, such as incorporating guidelines information. Typically, these methods do not involve training model parameters. (2) Code LLMs~\citep{Guo2023RetrievalAugmentedCG, Sainz2023GoLLIEAG, Bi2023CodeKGCCL, li2023codeie, wang2023code4struct}, which adopt the Code LLMs' capabilities of understanding structured information on IE tasks, often perform better than natural language LLMs. (3) Multi-task fine-tuning~\citep{lu2022unified, Wang2022DeepStruct, Wang2023InstructUIEMI, Sainz2023GoLLIEAG,chen2023one,zhou2023universalner}, which trains LLMs on multiple IE datasets, enhancing the models' performance on IE tasks, especially in cross-task scenarios. These works do not sufficiently align LLMs with IE tasks, due to the lack of diverse alignment data. These trained LLMs also exhibit a decline in general capabilities. In this paper, we aim to sufficiently align LLMs on IE tasks with rich alignment data without compromising their general capabilities.





\section{Alignment Data Construction}
\label{sec:alignment_Data_Construction}
This section introduces the construction process of \ourdata. The process mainly consists of $3$ steps: IE data collection (\cref{sec:ie_data_collect}), input construction (\cref{sec:input_org}), and answer generation (\cref{sec:answer_gen}).
Details of data construction are shown in appendix~\ref{sec:app_data_collection}.

\subsection{IE Data Collection}
\label{sec:ie_data_collect}



\begin{figure}
    \centering
    \includegraphics[width=0.9\linewidth]{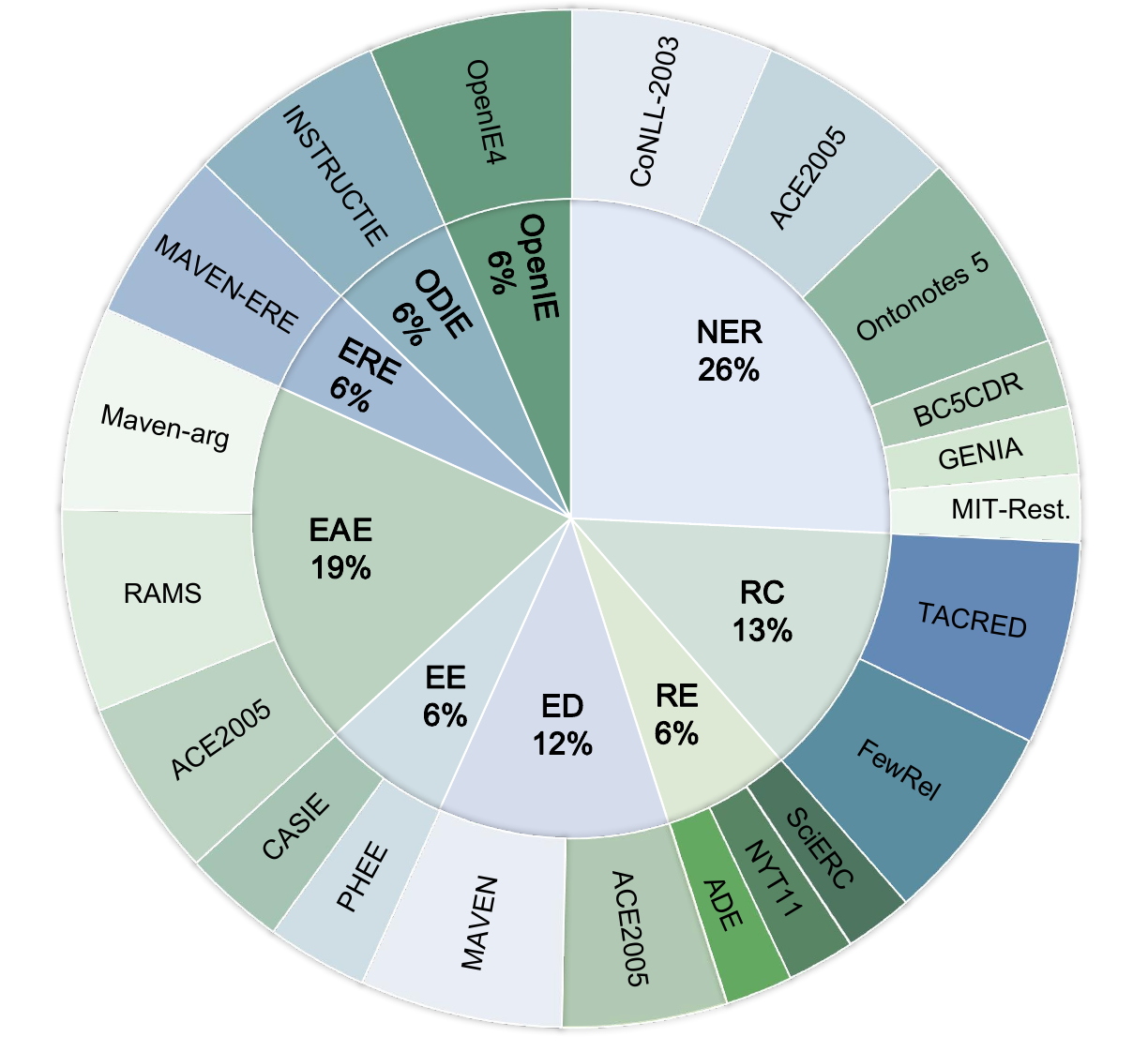}
    \caption{IE tasks, datasets, and respective proportions in \ourdata.}
    \label{fig:datasets}
\end{figure}

We first collect multiple IE datasets, including closed IE~\citep{Xu2023LargeLM}, open IE~\citep{liu2022open}, and on-demand IE~\citep{Jiao2023InstructAE}, covering various domains, such as general, financial, and biomedical domains. 
We filter out $80\%$ of NA data, which does not contain information needing extraction. To balance different datasets, we employ the examples-proportional mixture~\citep{wei2021finetuned}, with a dataset size limit of $5,000$. The data collection information is shown in Figure~\ref{fig:datasets}.


\begin{figure*}[t]
    \includegraphics[width=1.0\linewidth]{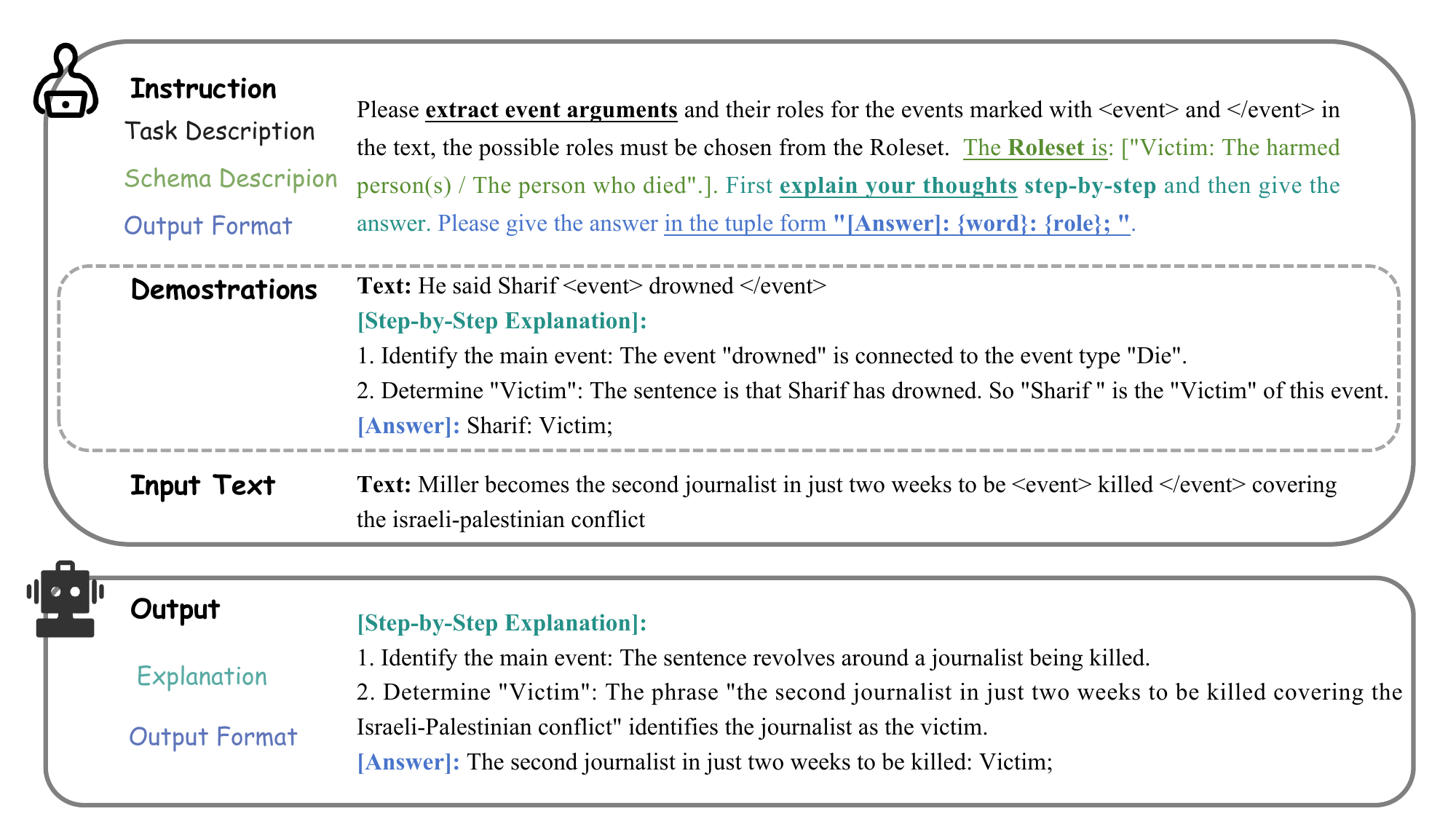} 
    \caption{
        An example of the input and output in \ourdata. $50\%$ of the data in \ourdata includes in-context demonstrations. The instruction consists of the descriptions of task, schema, and output format. The output consists of an explanation (for $10\%$ of the instances in \ourdata) and the answer adhering to the format in instruction.
    }
    \label{fig:example of input and output}
\end{figure*}
\subsection{Input Construction}
\label{sec:input_org}
We construct diverse input to better align LLMs on IE tasks. As shown in Figure~\ref{fig:example of input and output}, the input primarily consists of an instruction and a piece of input text. The instruction usually includes $3$ components: task description, schema description, and output format description. The schema description is only used in closed IE tasks, as open IE and on-demand IE do not include a schema.
Some inputs also include several demonstrations, i.e., input-output exemplars, for enhancing few-shot in-context learning capabilities. We introduce the augmentation process of the $3$ components of instructions and the construction of few-shot demonstrations.

\paragraph{Task Description}
For each IE task, we first manually craft $10$ task descriptions. Then we adopt GPT-3.5 to 
generate $20$ more descriptions. 
Specifically, to enrich the diversity of generated descriptions, similar to Self-Instruct, we employ an iterated generation process, which uses 
$3$ manually written descriptions and $2$ generated descriptions as the prompt for GPT-3.5 to generate a new description. 
Finally, we manually verify the generated descriptions and filter out those with hallucinations.

\paragraph{Schema Description}
For closed IE tasks, inspired by GoLLIE~\citep{Sainz2023GoLLIEAG}, we augment the schema descriptions, i.e., category information, from $3$ aspects: (1) Schema shuffling and sampling. We randomly shuffle the order of categories in the schema and select a random subset of $1$ to the maximum number of categories to include in the instruction. This technique aims to prevent model overfitting on the schemata in the training corpora, forcing the model to only output categories present in the input schema. 
(2) Incorporation of guidelines.
Guidelines are definitions of the schema, which can enhance the model's ability to understand the schema definition, thereby improving the model's zero-shot generalization capabilities on unseen tasks~\citep{Sainz2023GoLLIEAG}. Therefore, we add guidelines information to $20\%$ of the data in the training corpora. Similar to GoLLIE~\citep{Sainz2023GoLLIEAG}, we also include several examples for each category.
The remaining data does not include guidelines to prevent the model from memorizing schema definitions and to enhance data diversity.  (3) Replacing categories with symbols. We randomly replace category names with symbols (e.g., \texttt{LABEL\_1}) to prevent the model from overfitting to category names~\citep{Sainz2023GoLLIEAG} and enhance the in-context learning ability~\citep{wei2023symbol}.

\paragraph{Output Format Description}
LLMs sometimes struggle to follow the required output format in IE tasks~\citep{han2023information}. To enhance the model's ability to follow format requirements, we introduce various output format descriptions in the instructions, requiring the model to output accordingly. Specifically, for each closed IE and open IE task, there are mainly $3$ types of formats: (1) Triplet format, specifying output in various triple formats, e.g., \textit{(head entity; relation; tail entity)} or \textit{(head entity; tail entity; relation)} for relation extraction. (2) JSON format, requiring the model to output JSON formatted results. (3) Natural language format, without specific format requirements, allowing the model to output in natural language. The construction process of outputs corresponding to format requirements is detailed in \cref{sec:answer_gen}. On-demand IE does not involve output format descriptions, as its output is typically in a fixed Markdown format.

\paragraph{Few-shot Demonstrations}
Finally, to enhance the model's few-shot in-context learning capabilities, we augment the training corpus with few-shot demonstration inputs. Specifically, we randomly select $50\%$ of the training data and add $1$ to $8$ randomly sampled examplars to the original input. These examplars consist of a piece of input text and the output result, with the output format adhering to the requirements in the instruction. For each instance, the demonstrations are randomly sampled and shuffled to prevent the model from overfitting to fixed demonstrations.

\subsection{Answer Construction}
\label{sec:answer_gen}

We construct corresponding outputs according to the format requirements in the instructions generated in \cref{sec:input_org}. Specifically, for each closed IE and open IE task, the outputs include $3$ formats: (1) Triplet format. Following \citet{Wang2022DeepStruct}, we convert the output into serialized triplet form. For outputs containing multiple triplets, we randomly shuffle the order of triplets to mitigate potential order bias~\citep{li2023set}. (2) JSON format. We devise a set of JSON formats and transform the answers into corresponding JSON data. (3) Natural language format. We manually write several templates for natural language outputs for each task and construct corresponding outputs based on these templates. For on-demand IE, we adopt the original answers in their datasets~\citep{Jiao2023InstructAE}.

To enhance the model's intensive understanding of IE task procedures, we augment a subset ($10\%$) of instances with Chain-of-Thought (CoT)~\citep{wei2022chain} explanations for closed and open IE. 
To generate high-quality CoT explanations, we input both the input text and its ground truth answer to GPT-4.
Specifically, we sample $1,000$ instances for each task and then use the text input and its corresponding answer as inputs to generate CoT explanations. We randomly select $200$ instances to assess the quality of the CoT explanations and find that GPT-4 generally generates effective and informative step-by-step thoughts for the answers. 



\section{Model Training}
\label{sec:model_training}

This section introduces the alignment training process, including SFT~\citep{ouyang2022training} and DPO~\citep{dpo} training. 
More training details are placed in \cref{sec:app_training_details}.

For the SFT training, to preserve the model's general capabilities during alignment, we utilize the general alignment corpora used by TULU 2~\citep{Ivison2023CamelsIA}. Specifically, we \textbf{mix} \ourdata ($83,585$ instances) and $320,000$ instances of general alignment corpora as the training dataset. We adopt \textbf{LLAMA 2}~\citep{touvron2023llama} as the backbone model and train the model for $6,306$ gradient steps, resulting in \ourmodel.

After the SFT phase, we continue to train \ourmodel using the DPO objective. We first construct DPO training data, i.e., preference pairs (a preferred answer and a dispreferred answer). The original training objective of DPO requires online sampling of preference pairs from the model after SFT~\citep{dpo} with human annotation. In practice, some works also use human-annotated offline preference pairs for training, such as those sampled from other more powerful models~\citep{Ivison2023CamelsIA}. In our implementation, to obtain more diverse data, we used a mix of online and offline data. Unlike previous work where preference pairs need human annotation, there exists ground truth for IE and hence the preference pairs can be automatically constructed. 
Therefore, similar to~\citet{chen2024self}, we use the model itself outputs and original ground truths without needing extra human-annotated preference pairs, which is akin to self-improvement~\citep{huang2023large} and can sufficiently minimize manual involvement and conserves labors.
Specifically, we employ the BLEU~\citep{papineni2002bleu} score as the metric\footnote{We do not use the F1 score because some predictions are unstructured and we can not directly compute their F1 scores.} to automatically construct preference pairs. 
We sample the output of \ourmodel $5$ times for an instance with the sampling temperature as $1.0$. If the difference between the highest and lowest BLEU scores exceeds $10\%$, we treat the corresponding outputs as a preference pair, where the higher BLEU output is the preferred answer. We denote this data as online data. We also take the lowest BLEU output as the dispreferred answer and the ground truth as the preferred answer, and denote this data as offline data.
Finally, we create \ourdatadpo,  containing $3$k online preference pairs and $7$k offline preference pairs. Then, using the DPO objective, we train for additional $937$ gradient steps on \ourmodel to obtain \ourmodeldpo.





\section{Experiments}
\begin{table*}[ht]
\centering
\small{
\begin{tabular}{l|l|ccccc|c}
\toprule
\multirow{-2}{*}{}  & \multicolumn{1}{l|}{\multirow{1}{*}{Model}} & \multicolumn{1}{c}{FewNERD$_{\text{NER}}$} & \multicolumn{1}{c}{SemEval$_{\text{RC}}$} & \multicolumn{1}{c}{{\color[HTML]{333333} RichERE$_{\text{ED}}$}} & \multicolumn{1}{c}{RichERE$_{\text{EAE}}$} & \multicolumn{1}{c|}{MATRES$_{\text{ERE}}$} & \multicolumn{1}{c}{\multirow{1}{*}{AVG}} \\ \midrule
                            & GoLLIE                                        & $29.7$                                       & $29.2$                                      & $21.0$                                                             & $39.2$                                      & $25.9$                                       & $29.0$                                      \\
                            & InstructUIE                                 &   $33.5$                                    &   \boldsymbol{$43.9\dag$}                                  &    \underline{$40.8$}                                                         &    $17.4$                                   &  $30.2$                                     &     $33.2$                                      \\ \cmidrule{2-8} 
                            & \ourmodel                                     & $32.7$                                       & $21.8$                                      & $24.5$                                                             & $45.8$                                       & $47.8$                                       & $34.5$                                      \\
\multirow{-4}{*}{\rotatebox{90}{Zero-Shot}} & \ourmodeldpo                                      & $32.1$                                       & $22.9$                                      & $26.9$                                                             & $47.9$                                       & $47.9$                                       & $35.5$   \\ \midrule

                            & LLAMA 2                                    & {\color{white}0}$4.4$                                        & {\color{white}0}$8.2$                                       & {\color{white}0}$3.0$                                                              & {\color{white}0}$8.9$                                        & {\color{white}0}$3.8$                                        & {\color{white}0}$5.7$                                       \\
                            & TULU 2                                     & $24.4$                                       & $25.1$                                      & $11.8$                                                             & $24.4$                                       & $16.8$                                       & $20.5$                                      \\
                            & GoLLIE                                     & $30.0$                                       & $17.5$                                      & $19.1$                                                             & $24.3$                                       & $32.6$                                       & $24.7$                                      \\
                            & InstructUIE                               & $35.6$                                          & $38.3\dag$                                        & \boldsymbol{$42.7$}                                                                & $17.8$                                          & $10.4$                                          &  $29.0$                                         \\ 
                                                        & GPT-3.5*                                       & \underline{$44.1$}                                       & $24.0$                                      & $18.8$                                                             & $28.7$                                       & $41.0$                                       & $31.3$                                      \\
                            & GPT-4*                                         & \boldsymbol{$52.2$}                                       & \underline{$39.5$}                                      & $23.8$                                                             & $41.0$                                       & \boldsymbol{$59.0$}                                       & \boldsymbol{$43.1$}                                      \\ \cmidrule{2-8} 
                            & \ourmodel                                     & $39.0$                                       & $33.8$                                      & $38.1$                                                             & \boldsymbol{$54.2$}                                       & $48.0$                                       & $42.6$                                      \\
\multirow{-8}{*}{\rotatebox{90}{  Few-Shot}}  & \ourmodeldpo                                      & $37.9$                                      & $34.2$                                     & $39.7$                                                            & \underline{$53.5$}                                      & \underline{$48.1$}                                      & \underline{$42.7$}                                      
\\ \bottomrule
\end{tabular}
}
\caption{\label{table:performance_of_closedIE}F1 scores (\%) of investigated LLMs on held-out closed IE datasets. The highest scores are in \textbf{bold} and the second highest are \underline{underlined}. * means the scores of the models are sourced from~\citet{peng2023specification}. $\dag$~indicates that InstructUIE has been trained on the SemEval training set.}
\end{table*}

\subsection{Experimental Setup}

\paragraph{Baselines}
For closed IE, we primarily compare $3$ categories of models: (1) General open-source LLMs, including LLAMA 2~\citep{touvron2023llama}, a powerful foundation model and TULU 2~\citep{Ivison2023CamelsIA}, an instruction tuned LLAMA 2 model. We adopt the 7B version of these models. (2) Proprietary LLMs, including GPT-3.5~\citep{chatgpt} and GPT-4~\citep{openai2023gpt}. 
(3) Models optimized for IE tasks, including GoLLIE~\citep{Sainz2023GoLLIEAG},
a code LLM fine-tuned for IE tasks, 
and InstructUIE~\citep{Wang2023InstructUIEMI}, 
an LLM trained on multiple IE tasks. 
For open IE, we adopt the state-of-the-art model, OpenIE6~\citep{kolluru2020openie6}, as the baseline. For on-demand IE, we compare with the ODIE$_{\text{Direct}}$ model~\cite{Jiao2023InstructAE}, which is trained on on-demand IE training set.


\paragraph{Evaluation Datasets}
For closed IE and open IE, we utilize \textbf{held-out} datasets for evaluation, i.e., the datasets not included in the alignment corpora, to better assess the models' generalization capabilities on IE tasks. Specifically, for closed IE, we employ $4$ commonly used datasets: the NER dataset FewNERD~\citep{ding2021few}, the RC dataset SemEval~\citep{hendrickx2010semeval}, the ED and EAE dataset RichERE~\citep{song2015light}, and the ERE dataset MATRES~\citep{ning2018multi}. For open IE, we use the CaRB~\citep{bhardwaj2019carb} and ROBUST~\citep{qi2023preserving} datasets. For on-demand IE, we employ InstructIE~\citep{Jiao2023InstructAE}.


\paragraph{Evaluation Setup}
For closed IE and open IE, we adopt zero-shot and few-shot (4-shot for closed IE and 5-shot for open IE) in-context learning for evaluation. The few-shot demonstrations are randomly sampled from the corresponding training set. For on-demand IE, we adopt zero-shot evaluation the same as in the original paper~\citep{Jiao2023InstructAE}. For LLAMA 2, TULU 2, GoLLIE, and InstructUIE, we re-evaluate them using the same demonstrations. The results for GPT-3.5, GPT-4, OpenIE6, and ODIE$_{\text{Direct}}$ are obtained from previous work. 
Regarding evaluation metrics, we report F1 scores and employ the same calculation method as previous work. For details, please refer to \citet{peng2023specification} for closed IE, ~\citet{qi2023preserving} for open IE, and ~\citet{Jiao2023InstructAE} for on-demand IE. 
More evaluation details are placed in \cref{sec:app_experiment}.

\subsection{Experimental Results}

\begin{table}[t]
\small{
\resizebox{\linewidth}{!}{
\begin{tabular}{l|l|cc|c}
\toprule
\multicolumn{1}{l|}{}                          & \multicolumn{1}{l|}{Model}                        & \multicolumn{1}{c}{CaRB} & \multicolumn{1}{c|}{ROBUST} & \multicolumn{1}{c}{AVG}                     \\ \midrule
\multirow{2}{*}{Zero-Shot}                     & \ourmodel                                    &        $52.3$             &               $35.3$              &     $43.8$                                     \\
                                               & \ourmodeldpo                                     &       $53.0$              &               $36.6$             &     $44.8$ \\ \midrule
\multirow{5}{*}{Few-Shot}                      & LLAMA 2                                   & $10.9$                     & {\color{white}0}$0.2$                         & {\color{white}0}$5.6$                                      \\
                                               & TULU 2                                    & $32.5$                     & $11.0$                          & $21.8$                                     \\ 
                                               & GPT-3.5*                                      & $51.6$                     & $27.5$                        & $39.6$                                     \\ \cmidrule{2-5} 
                                               & \ourmodel                                    & \underline{$55.3$}                     & \underline{$38.5$}                        & \underline{$46.9$}                                     \\
                                               & \ourmodeldpo                                     & {\boldsymbol {$56.0$}}               & {\boldsymbol {$39.2$}}                  & {\boldsymbol {$47.6$}}                               \\ \midrule
Fine-Tuning                                      & OpenIE6*                                      & $55.2$            & $35.8$               & $45.5$                           
\\ \bottomrule
\end{tabular}
}
}
\caption{\label{table:performance_of_openIE}F1 scores (\%) of investigated LLMs on held-out open IE datasets. The highest scores are in \textbf{bold} and the second highest are \underline{underlined}. * denotes the results are obtained from \citet{qi2023preserving}.}
\end{table}
\paragraph{Results on Closed IE} 
The results on held-out closed IE datasets are shown in Table~\ref{table:performance_of_closedIE}. We can observe that: 
(1) \ourmodel performs significantly better than the original LLAMA 2 and surpasses all IE LLMs and GPT-3.5, on par with GPT-4. 
Compared to InstructUIE and GoLLIE, which adopt more advanced base LLMs (FLAN-T5 11B
and Code LLAMA 7B
) in IE tasks~\citep{peng2023specification} and more SFT data ($144$k and $165$k IE instances), \ourmodel achieves better results using only $83$k SFT data with LLAMA 2 7B.
This indicates that our data construction method is effective and \ourdata is of high quality.
(2) DPO further enhances performance. \ourmodeldpo performs consistently better than \ourmodel across most datasets. This suggests that for extractive tasks with ground truth answers, further alignment using DPO can also self-improve model performance. However, the improvement of DPO is generally modest, possibly due to not using additional human-annotated preference pairs. We leave using human-annotated preference pairs for training DPO as future work. 
(3) Incorporating in-context demonstrations during the alignment process is necessary. Previous work only focuses on zero-shot capabilities and overlooks few-shot capabilities of LLMs, resulting in no significant improvement or even a decline when providing few-shot demonstrations, e.g., a $4.3\%$ decline in F1 score for GoLLIE. In contrast, \ourmodel’s few-shot performance is much better than its zero-shot performance, which suggests that \ourmodel possesses few-shot in-context learning capabilities for closed IE tasks. It demonstrates
the effectiveness of including in-context demonstrations in the alignment process. 


\begin{table}[t]
\small{
\resizebox{\linewidth}{!}{
\begin{tabular}{l|cc|c}
\toprule
\multicolumn{1}{l|}{Model}                        & Table Header   & Table Content &         AVG             \\ \midrule
LLAMA 2                                   & $36.5$          & {\color{white}0}$8.2$                & $22.4$                 \\
TULU 2                                    & $66.9$          & $47.4$               & $57.2$                 \\ 
GPT-3.5*                                      & $\boldsymbol{74.5}$          & {\underline{$51.4$}}         & {\underline{$63.0$}}           \\
GPT-4*                                        & $\boldsymbol{74.5}$          & \boldsymbol{$59.1$}      & \boldsymbol{$66.8$}        \\ 
ODIE$_{\text{Direct}}$*                       & \underline{$73.8$}          & $45.9$               & $59.9$                 \\ \midrule
\ourmodel                                    & $73.4$ & $47.3$                 & $60.4$                 \\
\ourmodeldpo                                     & $73.7$   & $47.3$               & $60.5$                
\\ \bottomrule
\end{tabular}
}
}
\caption{\label{table:performance_of_ondemandIE}F1 scores (\%) of investigated LLMs on the on-demand IE task.  The highest scores are in \textbf{bold} and the second highest are \underline{underlined}. * means the scores of the models are sourced \citet{Jiao2023InstructAE}.}
\end{table}

\begin{figure}[t]
    \includegraphics[width=1.0\linewidth]{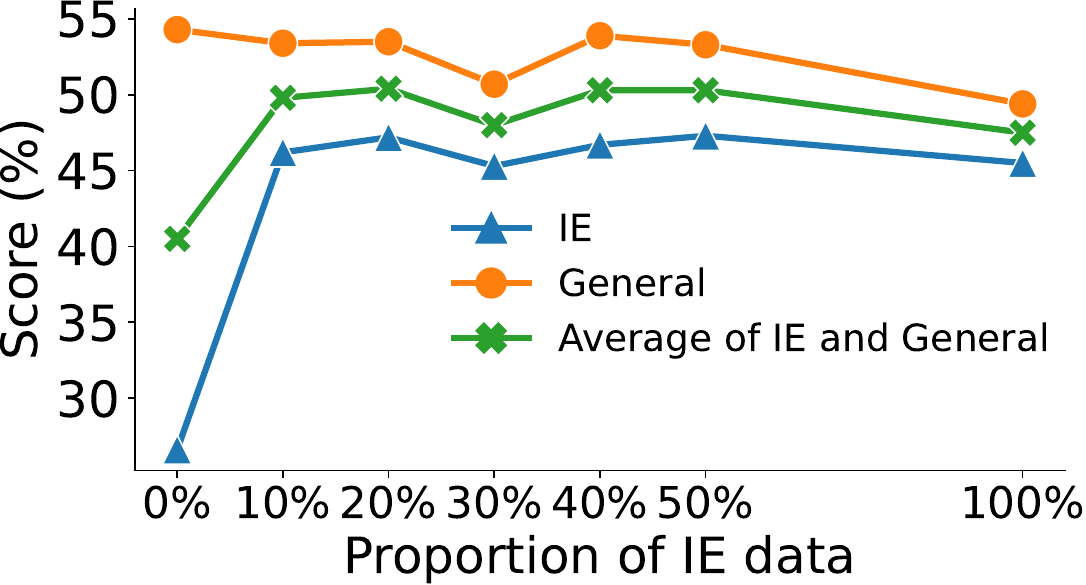} 
    \caption{
        Scores (\%) on IE tasks (average of closed IE, open IE, and on-demand IE) and general tasks (average of commonsense reasoning, MMLU, and BBH) of our model trained with varying proportions of IE data. We finally adopt a proportion of $20\%$ to train \ourmodel.
    }
    \label{fig:analysis_IE_rate}
\end{figure}

\paragraph{Results on Open IE}
The results on held-out open IE datasets are shown in Table~\ref{table:performance_of_openIE}.
The observations are similar to those in closed IE. \ourmodel and \ourmodeldpo perform much better than GPT-3.5, especially on ROBUST, a robust
open IE benchmark with ubiquitous syntactic transformations~\citep{qi2023preserving}, which demonstrates the robustness of our models on open IE. Our models even outperform the SoTA fine-tuned model, OpenIE6, demonstrating the effectiveness of alignment training.

\paragraph{Results on On-demand IE}
The results of the on-demand task are shown in Table~\ref{table:performance_of_ondemandIE}. 
On-demand IE uses two evaluation metrics: Table header, evaluating how well the model follows instructions, and table content, assessing the extraction quality~\citep{Jiao2023InstructAE}. We can observe that ADELIE achieves a competitive table header score to GPT-4, which suggests that ADELIE better understands and follows user instructions. It demonstrates that the alignment process effectively aligns ADELIE with user instructions and expectations.

In general, ADELIE achieves remarkable results across all IE tasks, particularly in few-shot evaluation scenarios, which demonstrates their strong zero-shot and few-shot generalization capabilities and the effectiveness of our alignment corpora \ourdata and \ourdatadpo.

\section{Analysis}
This section introduces further analyses of key factors in training the models (\cref{sec:analysis_general,sec:analysis_dpo}) and analyses on few-shot ICL capabilities (\cref{sec:analysis_fewshot}).

\subsection{Analysis on General Capabilities}
\label{sec:analysis_general}
\begin{table}
\centering
\small
\resizebox{\linewidth}{!}{
\begin{tabular}{l|cccc}
\toprule
\multicolumn{1}{c|}{\multirow{2}{*}{Model}} 
& \multicolumn{1}{c}{Commonsense} & \multicolumn{1}{c}{\multirow{2}{*}{MMLU}} & \multicolumn{1}{c}{\multirow{2}{*}{BBH}} 
& \multicolumn{1}{c}{\multirow{2}{*}{AVG}} 
\\
\multicolumn{1}{c|}{}                       
& \multicolumn{1}{c}{Reasoning}   
& \multicolumn{1}{c}{}                      
& \multicolumn{1}{c}{}                     
& \multicolumn{1}{c}{}\\ \midrule
FLAN-T5\textsubscript{11B} 
& 
$45.8$ & $32.1$ & \underline{$40.8$} 
&$43.7$\\
InstructUIE 
&$42.5$ &$30.4$ &$13.1$ 
&$37.9$\\
\midrule
LLAMA 2                                     
& $55.5$                            & $45.7$                                      & $35.7$      & $52.2$                               \\         
\quad \texttt{+General}
& \boldsymbol{$56.9$}                            & \boldsymbol{$49.3$}                                      & \boldsymbol{$41.7$}                                     
&\boldsymbol{$54.3$}\\
\ourmodel 
& $56.6$ & $47.1$ & $38.3$ 
&$53.5$\\
\ourmodeldpo 
& \underline{$56.8$} & \underline{$47.3$} & $38.9$
& \underline{$53.8$}
\\ \bottomrule
\end{tabular}
}
\caption{\label{tab:analysis_general}Performance (\%) on general benchmarks. ``\texttt{+General}'' is the model trained with only general alignment corpora for the same gradient steps as \ourmodel. InstructUIE is trained based on FLAN-T5\textsubscript{11B}.}
\end{table}

Alignment may impact the model's general capabilities, namely ``Alignment Tax''~\citep{bai2022training, kim2023aligning}. We investigate the general capabilities of previous LLMs for IE and \ourmodel in this section. Specifically, we select several widely-used benchmarks for assessing general capabilities: MMLU~\citep{Hendrycks2020MeasuringMM}, BBH~\citep{suzgun2023challenging}, and Commonsense Reasoning (including HellaSwag~\citep{zellers2019hellaswag}, WinoGrande~\citep{sakaguchi2021winogrande}, PIQA~\citep{bisk2020piqa}, SIQA~\citep{sap2019social}, ARC easy and challenge~\citep{clark2018think}, and OpenbookQA~\citep{mihaylov2018can}). The experimental details are placed in \cref{sec:app_general}.

Table~\ref{tab:analysis_general} presents the results. We can observe that: (1) InstructUIE suffers a significant decline in general capabilities compared to its original model, FLAN-T5\textsubscript{11B}~\citep{wei2021finetuned}, which indicates that using only IE data for alignment hurts the model's general capabilities. (2) \ourmodel's performance improves compared to the original LLAMA 2. Moreover, \ourmodel performs on par with the model trained specifically on general alignment data (\texttt{+General}). This suggests that mixing general and IE alignment data can both enhance the model's general and IE capabilities and hence mitigate the impact of ``Alignment Tax''. Therefore, we advocate for including \ourdata in the alignment data to enhance the model's capabilities.

We further investigate the impact of data mixing strategy. Specifically, we observe the performance of models trained with varying proportions of IE data from \ourdata in the overall alignment data. The results are shown in Figure~\ref{fig:analysis_IE_rate}. We can observe that: (1) There is a substantial improvement in IE tasks, even with only $10\%$ of the training data being IE data. This suggests a lack of IE data in the existing mainstream alignment data. 
(2) Adding IE data in training leads to a decrease in the model's general capabilities, but this decline is limited when the proportion is below $50\%$.
This may be due to the insufficient capacity of the 7B model, and we leave training a larger model as future work. 
Considering the results on both IE and general tasks, we ultimately train ADELIE on the data including $20\%$ IE data and $80\%$ general data.

\subsection{Analysis on DPO Training}
\label{sec:analysis_dpo}

\begin{figure}[t]
    \includegraphics[width=1.0\linewidth]{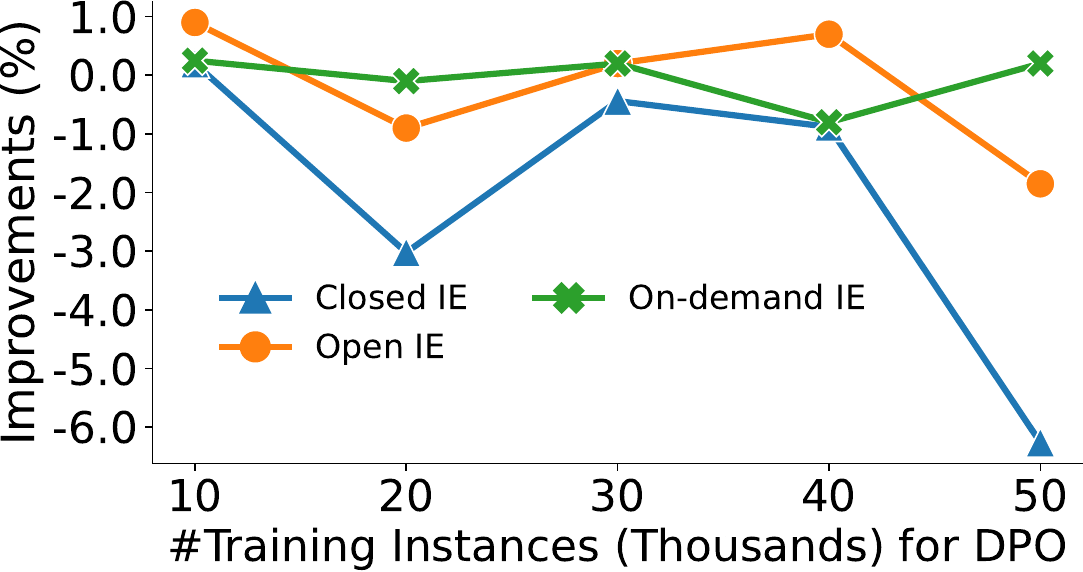} 
    \caption{Performance improvements (\%) of the model trained on varying scales of data, compared to \ourmodel before DPO training. 
    }
    \label{fig:Dpo_data_size}
\end{figure}

We analyze the training data construction strategy for DPO, i.e., the construction of preference pairs, each consisting of a preferred answer and a dispreferred answer. As mentioned in \cref{sec:model_training}, we adopt both offline and online data for training. The distinction lies in that both preferred and dispreferred answers of online data are sampled from \ourmodel's outputs, while the preferred answers of offline data are ground truths. 
We examine the impact of the proportion of offline data.
We find that generally the model trained on $70\%$ offline data and $30\%$ online data performs best, with an average $47.7\%$ F1 score across closed, open, and on-demand IE tasks. 
The detailed results are shown in Appendix~\ref{sec:app_dpo}. We also explore the impact of data size on performance, as shown in Figure~\ref{fig:Dpo_data_size}. We find that $10$k instances is sufficient to train the model, and using more data increases computational costs without significant improvements. 
This may be due to not using additional human-annotated data, leading to model overfitting.
Therefore, \ourdatadpo ultimately consists of $2,996$ online and $6,989$ offline instances. 


\begin{figure}
    \centering
\includegraphics[width=1.0\linewidth]{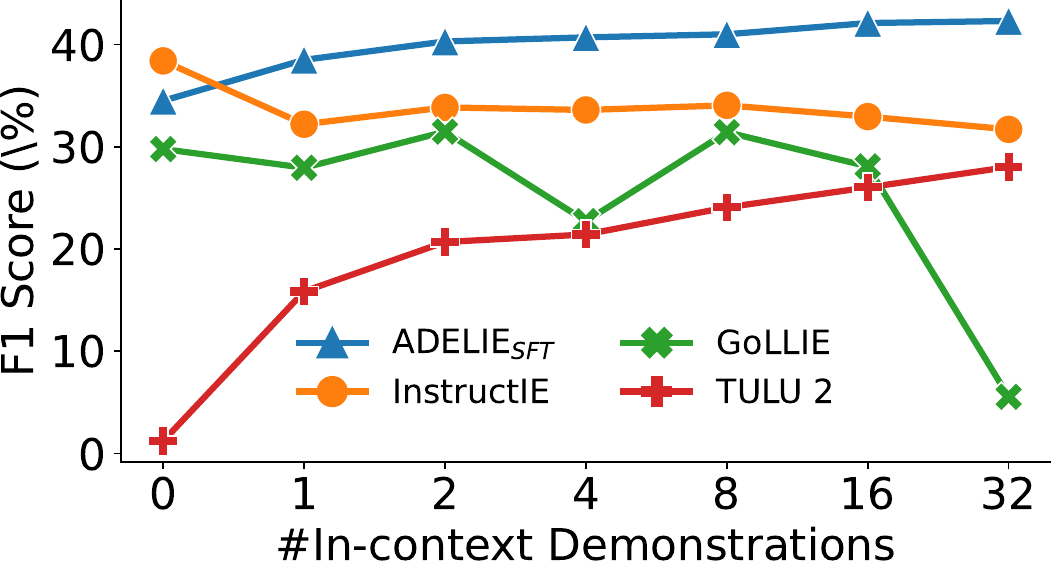}
    \caption{F1 scores (\%) using a varying number of in-context demonstrations on closed IE, excluding MATRES (document-level) due to the limited context size. }
    \label{fig:few_shot}
\end{figure}

\subsection{Analysis on Few-shot ICL Capabilities}
\label{sec:analysis_fewshot}
Closed IE typically includes a schema with multiple predefined categories and hence needs more in-context demonstrations to effectively illustrate these categories~\citep{li2024long}, which necessitates the few-shot in-context learning (ICL) capabilities of the model. We observe \ourmodel's few-shot ICL capabilities, as presented in Figure~\ref{fig:few_shot}. We find that \ourmodel performs consistently better with more demonstrations, even though 
\ourmodel is trained with a maximum of only $8$ demonstrations. In contrast, InstructUIE and GoLLIE suffer a decline with more few-shot demonstrations. This demonstrates the effectiveness of using in-context demonstrations during the alignment process.



\section{Conclusion}
This work introduces ADELIE, a series of LLMs aligned for information extraction tasks. ADELIE includes \ourmodel, which is supervised fine-tuned on \ourdata with high-quality $83,585$ instances, and \ourmodeldpo, which further trains \ourmodel on $9,985$ preference pairs (\ourdatadpo) using DPO. Extensive experiments demonstrate that ADELIE achieves impressive results on IE tasks, particularly in the few-shot setting. We hope our work can provide meaningful insights for future model alignment efforts.


\section*{Acknowledgements}
We thank all the anonymous reviewers and meta reviewers for their valuable comments. This work is supported by the National Natural Science Foundation of China (No. 62277033), Beijing Natural Science Foundation (L243006), a grant from the Institute for Guo Qiang, Tsinghua University (2019GQB0003) and the project from Tsinghua-SPD Bank Joint-Lab. Thanks the support from National Engineering Laboratory for Cyberlearning and Intelligent Technology, and Beijing Key Lab of Networked Multimedia.

\section*{Limitations}

The limitations of this work are mainly threefold: 
(1) The preference pairs used for DPO training are automatically constructed without additional human annotation, which may limit the performance of DPO-trained models. We leave using human-annotated preference pairs for DPO training as the future work.
(2) We train only with a 7B scale model due to computational limits. Employing a larger-scale model can yield better performance, but it does not impact the conclusions of this paper.
(3) This paper only involves English data. In the future, we will try to support more languages, and we encourage researchers to explore aligning models for multilingual information extraction.

\section*{Ethical Considerations}
We discuss potential ethical concerns of this work: 
(1) \textbf{Intellectual property}. Our work utilizes multiple widely-used IE datasets, and we strictly adhere to the licenses of these datasets. We will share \ourdata and \ourdatadpo the CC BY-SA 4.0 license\footnote{\url{https://creativecommons.org/licenses/by-sa/4.0/}}. \ourdata and \ourdatadpo include some data only accessible to Linguistic Data Consortium\footnote{\url{https://www.ldc.upenn.edu/}} (LDC) members, e.g., ACE 2005~\citep{ace2005}. 
For these parts, we will release only the data processing scripts.
(2) \textbf{Intended use}. This paper introduces ADELIE, aiming to align LLMs and enhance their performance on IE tasks.
(3) \textbf{Potential risk control}. 
\ourdata and \ourdatadpo are collected and constructed based on widely-used public data and data obtained from GPT-3.5 and GPT-4. We believe that these data have been well anonymized and sanitized by their original publishers and OpenAI. We also randomly sampled $100$ instances and found no sensitive data. 
(4) \textbf{AI assistance}. We adopt GPT-4 for paraphrasing some sentences when writing this paper.


\bibliography{custom}

\appendix
\clearpage
\section*{Appendices}
\section{Data Collection}
\label{sec:app_data_collection}
This section introduces details on data construction of \ourdata, including details of Input Construction (appendix~\ref{sec:app_input_construct}) and Answer Construction (appendix~\ref{sec:app_answer_construct}). In the data construction phase, we utilized \texttt{gpt-3.5-turbo-1106} for GPT-3.5 and \texttt{gpt-4-0125-preview} for GPT-4. 
The temperature parameter was set at $0.7$, with all other parameters at their default settings.

\subsection{Input Construction}
\label{sec:app_input_construct}

\begin{table*}[!htbp]
    \scalebox{0.9}{
    \begin{tabularx}{\textwidth}{X}
        \toprule
        \textbf{Prompt} \\
        You need to follow the template list to come up with a set of diverse templates. \\
        The task indicated by this template is the “Open Information Extraction” task.\\
        We need to write the instruction, input format and corresponding output format template for it.\\
        Instruction is an introduction to OpenIE tasks.\\
        The instruction template content should include the following strings to facilitate subsequent replacement of the content: \{text\}. \\
        The answer template content should include the following strings to facilitate subsequent replacement of the content: \{subject\}, \{predicate\}, \{object\}, \{time\}, \{location\}.\\
        Here are the requirements:\\
        1. Try not to repeat the verb for each template to maximize diversity.\\
        2. The language used for the template also should be diverse. For example, use interrogative sentences, imperative sentences, etc. \\
        3. Input and output templates ([Answer]: ..) should also be as diverse as possible.\\
        4. Do not repeat the format of the answer template, nor repeat the examples given.\\
        5. Input and output must correspond to each other.\\
        6. The templates should be in English.\\
        \\
        Template 1: \\
            (1) Instruction: Open information extraction requires the extraction of all relations in the sentence, i.e., predicates, the subjects and objects corresponding to these relations, and the possible time and place thesis elements. Please extract all the relational tuples in the following sentence: \{text\}. If there are no relations in the text, please answer "NA". \\
            (2) Fail output: NA. \\
            (3) Input template: Please give the answer in the tuple form "[Answer]: (\{predicate\}; \{subject\}; \{object\}; \{time\}; \{location\})". \\
            (4) Answer template: (\{predicate\}; \{subject\}; \{object\}; \{time\}; \{location\}), \\

        ...
        \\
        Please follow the format given in the example to generate 1 templates. \\
       \midrule
       \textbf{Response} \\
       Template 6: \\
            (1) Instruction: In the task of Open Information Extraction, your goal is to dissect the given text to unearth the underlying relationships. This involves pinpointing the subjects, predicates, and objects, as well as noting any relevant temporal or spatial details. Should the text lack relational tuples, kindly respond with 'No relationships identified.'. Text: \{text\}.  \\
            (2) Fail output: No relationships identified.\\ 
            (3) Input template: What connections can you draw between the subject and object, including any pertinent temporal or spatial details? \\
            (4) Answer template: Between "\{subject\}" and "\{object\}", the connection "\{predicate\}" is established, occurring at "\{time\}" and within "\{location\}". \\
       \bottomrule
    \end{tabularx}
    }
    \caption{\label{table:prompt4selfinstruct} An example prompt for Open IE to generate task description and output format description.}
\end{table*}



\begin{table*}[!htbp]
    \scalebox{0.9}{
    \begin{tabularx}{\textwidth}{X}
        \toprule
        \textbf{Prompt} \\
        Please generate a step-by-step explanation for $[$Answer$]$ based on $[$Question$]$, and give reasons for each step. \\
        The generated explanation should make use of the content in the $[$Question$]$ as much as possible, and must be consistent with the $[$Answer$]$.  \\
        It will eventually be provided at the front of the answer. \\
        No more than \{words\_number\} words.\\ \\
        $[$Question$]$: \{input\} \\
        $[$Answer$]$: \{output\} \\
        $[$Step-by-Step Explanation$]$:  \\
       \bottomrule
    \end{tabularx}
    }
    \caption{\label{table:prompt4explanation} A prompt template for generating explanations of answers, where placeholders \{words\_number\}, \{input\}, and \{output\} are replaced during usage.}
\end{table*}

\begin{table*}[ht]
\small{
\resizebox{\linewidth}{!}{
\begin{tabular}{l|lll|r|c|c}
\toprule
\multicolumn{1}{l|}{Tasks} & Datasets & Domain & \#Instances & \multicolumn{1}{l|}{\#total} & \multicolumn{1}{l|}{\#task desc.} & \multicolumn{1}{l}{\#output format desc.} \\ \midrule
 & CoNLL-2003~\cite{sang2003introduction} & General & 5000 &  &  &  \\
 & ACE2005$_{\text{NER}}$~\cite{ace2005} & General & 5000 &  &  &  \\
 & Ontonotes 5~\cite{pradhan2013towards} & General & 5000 &  &  &  \\
 & BC5CDR~\cite{li2016biocreative} & Biomedical & 1666 &  &  &  \\ 
 & GENIA~\cite{kim2003genia} & Biomedical & 1667 &  &  &  \\ 
\multirow{-6}{*}{NER} & {\color[HTML]{000000} MIT-Restaurant~\cite{liu2013asgard}} & Queries & 1667 & \multirow{-6}{*}{20000} & \multirow{-6}{*}{31} & \multirow{-6}{*}{15} \\ \midrule
 & TACRED~\cite{zhang2017position} & General & 5000 &  &  &  \\
\multirow{-2}{*}{RC} & FewRel~\cite{han2018fewrel} & General & 5000 & \multirow{-2}{*}{10000} & \multirow{-2}{*}{31} & \multirow{-2}{*}{15} \\ \midrule
 & SciERC~\cite{luan2018multi} & Scientific & 3332 &  &  &  \\
 & NYT11~\cite{takanobu2019hierarchical} & News & 3334 &  &  &  \\
\multirow{-3}{*}{RE} & ADE~\cite{GURULINGAPPA2012885} & Biomedical & 3334 & \multirow{-3}{*}{10000} & \multirow{-3}{*}{31} & \multirow{-3}{*}{15} \\ \midrule
 & ACE2005$_{\text{ED}}$~\cite{ace2005} & General & 4067 &  &  &  \\
\multirow{-2}{*}{ED} & MAVEN~\cite{wang2020maven} & General & 5000 & \multirow{-2}{*}{9067} & \multirow{-2}{*}{35} & \multirow{-2}{*}{15} \\ \midrule
 & PHEE~\cite{sun2022phee} & Biomedical & 2500 &  &  &  \\
\multirow{-2}{*}{EE} & CASIE~\cite{satyapanich2020casie} & Cybersecurity & 2500 & \multirow{-2}{*}{5000} & \multirow{-2}{*}{35} & \multirow{-2}{*}{15} \\ \midrule
 & ACE2005$_{\text{EAE}}$~\cite{ace2005} & General & 4420 &  &  &  \\
 & RAMS~\cite{li2021document} & General & 5000 &  &  &  \\
\multirow{-3}{*}{EAE} & Maven-arg~\cite{wang2023maven} & General & 5000 & \multirow{-3}{*}{14420} & \multirow{-3}{*}{27} & \multirow{-3}{*}{15} \\ \midrule
\multicolumn{1}{l|}{ERE} & MAVEN-ERE~\cite{wang2022maven} & General & 4278 & \multicolumn{1}{c|}{4278} & \multicolumn{1}{c|}{30} & \multicolumn{1}{c}{15} \\ \midrule
\multicolumn{1}{l|}{OpenIE} & OpenIE6~\cite{kolluru2020openie6} & General & 5000 & \multicolumn{1}{c|}{5000} & \multicolumn{1}{c|}{17} & \multicolumn{1}{c}{15} \\ \midrule
\multicolumn{1}{l|}{ODIE} & INSTRUCTIE~\cite{Jiao2023InstructAE} & General & 4904 & \multicolumn{1}{c|}{4904} & \multicolumn{1}{c|}{-} & \multicolumn{1}{c}{-} 
\\ \bottomrule
\end{tabular}
}
}
\caption{Training Datasets for the \ourdata dataset. }
\label{table:datasets}
\end{table*}

\begin{table}[ht]
\centering
\small{
\begin{tabular}{l|lcc}
\toprule
\multicolumn{1}{l|}{Tasks} & Datasets & \#Instances & $\Delta$ \\ \midrule
\multirow{3}{*}{NER} & CoNLL-2003 & 883 & 0.74 \\
 & ACE2005 & 854 & 0.79 \\
 & Ontonotes 5 & 855 & 0.82 \\ \midrule
\multirow{2}{*}{RC} & TACRED & 812 & 0.83 \\
 & FewRel & 733 & 0.84 \\ \midrule
\multirow{2}{*}{ED} & ACE2005 & 753 & 0.83 \\
 & MAVEN & 770 & 0.78 \\ \midrule
\multirow{3}{*}{EAE} & ACE2005 & 810 & 0.79 \\
 & RAMS & 767 & 0.78 \\
 & Maven-arg & 851 & 0.71 \\ \midrule
\multicolumn{1}{l|}{ERE} & MAVEN-ERE & 541 & 0.92 \\ \midrule
\multicolumn{1}{l|}{ODIE} & INSTRUCTIE & 617 & 0.87 \\ \midrule
\multicolumn{1}{l|}{OpenIE} & OpenIE4 & 739 & 0.81 \\
\bottomrule
\end{tabular}
}
\caption{Detailed information for the \ourdatadpo dataset. $\Delta$ represents the average difference in scores between the preferred and dispreferred answers in each dataset, with the score of the ground truth set to $1$.}
\label{tab:dpo datasets}
\end{table}

For constructing the Task Description and Output Format Description, we initially manually wrote $10$ task descriptions and $5$ output format descriptions for each task. We employed GPT-3.5 to generate task descriptions with the same semantics but varied expressions, as well as diverse output formats.

Table~\ref{table:prompt4selfinstruct} is an example used in the open IE task. Each generation includes five components: (1) Instruction: a description of the task. (2) Fail output: a response for when the task fails, which should correspond to the final requirement of the instruction. (3) Input template: a description of the output format in natural language, which must include multiple forms, such as triplets or natural language. (4) Output template: the output format corresponding to the input template.
Table~\ref{table:datasets} details the number of augmented descriptions generated for each task.

\subsection{Answer Construction}
\label{sec:app_answer_construct}


We employed GPT-4 to generate Chain-of-Thought explanations. 
Figure \ref{fig:example of input and output} illustrates examples of the explanations produced. 
We generate questions based on the Prompt template in Table~\ref{table:prompt4explanation}.
Moreover, to enhance diversity, we imposed a length constraint on the generated explanations, setting limits randomly between $70$ and $200$ words.


\section{Training Details}
\label{sec:app_training_details}
This section introduces the training data details (appendix~\ref{sec:app_training_datasets}), and training hyper-parameters (appendix~\ref{sec:app_training_hyperparameters}).
We performed each experiment once.

\subsection{Datasets details}
\label{sec:app_training_datasets}

\paragraph{\ourdata}
The process of constructing \ourdata involves the following steps: 
we sampled $5,000$ instances from these raw datasets.
Then, we followed the process outlined in \cref{sec:alignment_Data_Construction} and filtered out instances longer than $2,048$ tokens to prevent them from affecting the training effectiveness.

Finally, we compiled the \ourdata dataset, which consists of a total of $83,585$ high-quality IE data instances.
Table~\ref{table:datasets} shows the number of instances for each training dataset.

\paragraph{\ourdatadpo}


To generate \ourdatadpo, we sampled $50,000$ entries from raw datasets for processing in a manner similar to \ourdata. 
The sole distinction lies in the consistency of the output format with that required by the evaluation datasets, as shown in \cref{sec:app_evaluation_details}, aimed at facilitating more accurate BLEU scoring.
For calculating BLEU scores, we used \texttt{sentence\_bleu} function from \texttt{nltk.translate.bleu\_score}, with \texttt{SmoothingFunction} set to \texttt{method3}. 
Table~\ref{tab:dpo datasets} displays the information for the \ourdatadpo dataset, which consists of a total of $9,985$ instances.

\subsection{Training Hyperparameters}
\label{sec:app_training_hyperparameters}

\paragraph{SFT training}
To train the models, we employ supervised fine-tuning, which is the most effective method for aligning the models. 
The models were trained for $2$ epochs with an effective batch-size of $128$, a learning-rate of $2e-5$ with cosine scheduler and a warm-up phase of $0.03$.
To better facilitate learning in few-shot settings and document-level information extraction, the context length is set to $2048$ tokens.
we conduct SFT on Nvidia A100 GPUs, totaling approximately $120$ GPU hours.

\paragraph{DPO training}
Similar to SFT, we train the DPO model for $3$ epochs. 
Model is trained with a global batch size of $32$. 
And we employ a linear learning rate scheduler with a peak learning rate of $5e-7$ and a $0.1$ warm-up phase. 
The final \ourmodeldpo model is initialized from the SFT model, which was trained for $2$ epochs and further optimized for $3$ DPO epochs.
we conduct DPO on Nvidia A100 GPUs, totaling approximately $8$ GPU hours.

\section{Experimental Details}
\label{sec:app_experiment}
This section introduces the details of the experiment, including the details of the evaluation(appendix~\ref{sec:app_evaluation_details}), the inference details of the comparison baseline (appendix~\ref{sec:app_implementation_details}), and the detail results about analytical experiments (appendix~\ref{sec:app_general}, appendix~\ref{sec:app_dpo}).

\subsection{Evaluation Details}
\label{sec:app_evaluation_details}
\begin{table*}[!htbp]
    \scalebox{0.9}{
    \begin{tabularx}{\textwidth}{X}
        \toprule
        \textbf{[NER]} \\
        Please give the answer in the form "[Answer]: \{entity\}: \{type\}; ". \\
        \textbf{[RC]} \\
        Please give the answer in the tuple form "[Answer]: (\{subject\}; \{relation\}; \{object\}); ". \\
        \textbf{[ED]} \\
        Please give the answer in the form "[Answer]: \{event\}: \{class\}; ". \\
        \textbf{[EAE]} \\
        Please give the answer in the form "[Answer]: \{word\}: \{role\}; ". \\
        \textbf{[ERE]} \\
        Please give the answer in the tuple form "[Answer]: (\{first event\}; \{relation\}; \{second event\}); ". \\
         \textbf{[Open IE]} \\
        Please give the answer in the tuple form "[Answer]: (\{predicate\}; \{subject\}; \{object\}; \{time\}; \{location\})". If one or more of the last three elements does not exist, it can be omitted. \\
       \bottomrule
    \end{tabularx}
    }
    \caption{\label{tab:testset_foramt} The output format description for the hold-out tasks.}
\end{table*}

During the inference stage, we set the temperature to $0.01$ to ensure reproducible results.

\paragraph{Evaluation Input Construction}
The input composition of the evaluation test dataset is consistent with the training set, as shown in Figure~\ref{fig:example of input and output}. The only difference is that the output format description for each task is singular to facilitate automated evaluation.
Table~\ref{tab:testset_foramt} details the output format descriptions used for each task.
\paragraph{Evaluation Metrics}
In the closed IE tasks, we utilized exact matching to calculate the F1 score. 
For the open IE tasks on two benchmarks, we employed the same F1 calculation method as used by~\citet{qi2023preserving}. 
In on-demand IE tasks, following~\citet{Jiao2023InstructAE}, we adopted a soft matching strategy for assessing table headers and used the ROUGE-L F1 score to evaluate table content.

\subsection{Inference Details}
\label{sec:app_implementation_details}
We present the inference details of each baseline comparison.
(1) For general open-source LLMs, including LLAMA 2 7B (\texttt{meta-llama/Llama-2-7b}\footnote{https://huggingface.co/meta-llama/Llama-2-7b}) and TULU 2 7B (\texttt{allenai/tulu-2-7b}\footnote{https://huggingface.co/allenai/tulu-2-7b}). The test set and the prompts used for testing are completely consistent with \ourmodel.
(2) For models optimized for IE tasks, including GoLLIE 7B (\texttt{HiTZ/GoLLIE-7B}\footnote{https://huggingface.co/HiTZ/GoLLIE-7B} and InstructUIE (\texttt{ZWK/InstructUIE}\footnote{https://huggingface.co/ZWK/InstructUIE}. We observed that these models are sensitive to prompts, and directly using the testing prompts from \ourmodel 
leads to a sharp decline in model performance. Therefore, while keeping the test data unchanged, we adjusted the prompts to match the official formats of these models. For GoLLIE, as it did not provide formats for ERE and RC tasks, We modified the format of the RE task for adaptation purposes.

\subsection{Analysis on General Capabilities}
\begin{table*}
\small{
\resizebox{\linewidth}{!}{
    \begin{tabular}{l|ccccccccc|c}
    \toprule
    Model & MMLU & BBH & \multicolumn{1}{c}{HellaSwag} & \multicolumn{1}{c}{ARC easy} & \multicolumn{1}{c}{ARC challenge} & \multicolumn{1}{c}{WinoGrande} & \multicolumn{1}{c}{OpenbookQA} & \multicolumn{1}{c}{SIQA} & \multicolumn{1}{c|}{PIQA} & AVG \\ \midrule
    \ourmodel & 47.1 & 38.3 & 57.3 & 78.6 & 46.9 & 69.3 & 32.8 & 32.9 & 78.5 & 53.5 \\
    \ourmodeldpo & 47.3 & 38.8 & 57.5 & 78.9 & 47.3 & 69.2 & 33.0 & 33.1 & 78.8 & 53.8 \\ \midrule
    LLAMA 2 & 45.7 & 35.7 & 57.1 & 76.3 & 43.3 & 69.1 & 31.6 & 32.9 & 77.9 & 52.2 \\
    \multicolumn{1}{c|}{\quad \texttt{+General}} & 49.3 & 41.7 & 57.9 & 78.7 & 47.4 & 69.4 & 33.0 & 32.8 & 78.8 & 54.3 \\ \midrule
    FLAN-T5\textsubscript{11B} & 32.1 & 40.8 & 46.4 & 62.4 & 34.7 & 54.7 & 19.2 & 31.8 & 71.3 & 43.7 \\
    InstructUIE & 30.4 & 13.1 & 39.6 & 58.0 & 31.0 & 50.9 & 17.8 & 33.4 & 66.7 & 37.9 \\ \midrule
    IE Proportion=0.1 & 46.3 & 41.1 & 57.5 & 76.8 & 45.4 & 70.2 & 32.6 & 33.0 & 78.1 & 53.4 \\
    IE Proportion=0.3 & 31.7 & 38.3 & 55.1 & 75.7 & 43.9 & 69.9 & 31.2 & 32.9 & 78.1 & 50.7 \\
    IE Proportion=0.4 & 47.3 & 39.0 & 57.7 & 79.4 & 47.8 & 69.1 & 32.8 & 33.4 & 78.3 & 53.9 \\
    IE Proportion=0.5 & 47.7 & 39.0 & 57.8 & 77.6 & 44.4 & 70.6 & 31.0 & 33.5 & 78.2 & 53.3 \\
    IE Proportion=1.0 & 38.9 & 23.2 & 56.5 & 74.1 & 40.0 & 69.5 & 31.6 & 32.9 & 77.5 & 49.4 \\
    \bottomrule
    \end{tabular}
}
}
\caption{The performance of the models on general tasks in the analysis study for general capabilities.}
\label{tab:performance_detail_general_tasks}
\end{table*}

\begin{table*}
\small{
\resizebox{\textwidth}{!}{
\begin{tabular}{r|rccccccccc|c}
\toprule
 & Models & FewNERD$_{\text{NER}}$ & SemEval$_{\text{RC}}$ & RichERE$_{\text{ED}}$ & RichERE$_{\text{EAE}}$ & MATRES$_{\text{ERE}}$ & CaRB & ROBUST & Table Header & Table Content & AVG \\ \midrule
 & \ourmodel & 39.0 & 33.8 & 38.1 & 54.2 & 48.0 & 55.3 & 38.5 & 73.4 & 47.3 & 47.5 \\ \midrule
 \multirow{4}{*}{\rotatebox{90}{\#Training}} & 10k & 37.9 & 34.2 & 39.7 & 53.5 & 48.1 & 56.0 & 39.2 & 73.7 & 47.3 & 47.7 \\
 & 20k & 34.9 & 36.2 & 33.0 & 46.4 & 47.4 & 54.3 & 37.3 & 73.3 & 47.2 & 45.6 \\
 & 30k & 36.9 & 34.9 & 38.3 & 53.4 & 47.3 & 55.4 & 38.4 & 73.7 & 47.4 & 47.3 \\
 & 40k & 36.6 & 34.3 & 38.8 & 52.2 & 46.7 & 55.8 & 39.0 & 72.5 & 46.6 & 46.9 \\ \midrule
\multirow{7}{*}{\rotatebox{90}{Offline Data Rate}} & 0.0 & 38.7 & 34.1 & 37.6 & 54.1 & 46.9 & 55.1 & 38.0 & 73.8 & 47.3 & 47.3 \\
 & 0.3 & 38.4 & 33.2 & 39.8 & 53.8 & 47.3 & 55.2 & 38.2 & 73.7 & 47.5 & 47.5 \\
 & 0.5 & 38.6 & 34.2 & 40.7 & 53.8 & 47.5 & 55.2 & 38.2 & 73.4 & 47.1 & 47.6 \\
 & 0.6 & 38.2 & 33.9 & 38.6 & 53.7 & 47.3 & 55.4 & 38.5 & 73.8 & 47.6 & 47.4 \\
 & 0.7 & 37.9 & 34.2 & 39.7 & 53.5 & 48.1 & 56.0 & 39.2 & 73.7 & 47.3 & 47.7 \\
 & 0.8 & 37.9 & 34.4 & 39.2 & 53.8 & 47.8 & 55.6 & 38.8 & 73.6 & 46.9 & 47.6 \\
 & 1.0 & 37.8 & 35.1 & 40.0 & 53.7 & 47.7 & 55.5 & 38.8 & 73.7 & 46.9 & 47.7 \\
 \bottomrule
\end{tabular}
}
}
\caption{The performance of models in the DPO training analysis experiment across various IE tasks. The phrase "Training Offline" denotes maintaining data proportions at $0.7$ across different DPO training sets. "Offline Data Rate" refers to the proportion of offline data when the training set size is $10$k.}
\label{tab:app_dpo}
\end{table*}
\label{sec:app_general}
For the MMLU task, we conducted testing using $5$-shot. For the BBH task, we conducted testing using $3$-shot with COT. For the remaining commonsense reasoning tasks, we employed a uniform $0$-shot approach.
Table~\ref{tab:performance_detail_general_tasks} presents test results for detail.

\subsection{Analysis on DPO Training}
\label{sec:app_dpo}
Table~\ref{tab:app_dpo} presents the results in the DPO training analysis experiment.
We observed a trend in which the average performance initially increased and then decreased with the increase in the offline rate. The highest performance was achieved at $0.7$, reaching $47.73\%$ (although the result displayed for $1.0$ is also $47.7\%$, it is actually $47.68\%$).

\end{document}